\documentclass[letterpaper, 10 pt, conference]{ieeeconf}  

\IEEEoverridecommandlockouts                              
\usepackage[para,online,flushleft]{threeparttable}
\usepackage{booktabs}
\usepackage{multirow}
\usepackage{adjustbox}
\usepackage{graphicx}
\usepackage{siunitx}
\usepackage[utf8]{inputenc}
\usepackage[T1]{fontenc}
\usepackage{amsfonts}
\usepackage{textcomp, gensymb}
\usepackage{pifont}
\let\labelindent\relax
\usepackage{enumitem}

\usepackage{amsthm}

\usepackage{epsfig} 
\usepackage{amsmath} 
\usepackage{amssymb}  
\usepackage{siunitx}
\usepackage{subcaption}
\usepackage{caption}

\usepackage{color}
\usepackage[table]{xcolor}

\usepackage{soul}
\usepackage{theoremref}
\usepackage[noadjust]{cite}
\usepackage{amsthm}
\usepackage{mathtools}
\usepackage[linesnumbered,ruled,vlined]{algorithm2e}
\usepackage[symbol]{footmisc}

\newtheorem{theorem}{Theorem}
\newtheorem{lemma}[theorem]{Lemma}
\newtheorem{proposition}[theorem]{Proposition}
\newtheorem{corollary}{Corollary}[theorem]

\theoremstyle{remark}
\newtheorem*{remark}{Remark}

\usepackage{hyperref}
\hypersetup{
    colorlinks=true,
    linkcolor=blue,
    filecolor=magenta,      
    urlcolor=blue,
    pdftitle={Overleaf Example},
    pdfpagemode=FullScreen,
    }

\usepackage[acronyms, shortcuts]{glossaries}
\usepackage[noabbrev,capitalise]{cleveref}

\newtheorem*{theorem*}{Theorem}

\title{\LARGE\bf Generalized Benders Decomposition with Continual Learning for Hybrid Model Predictive Control in Dynamic Environment}

\author{Xuan Lin$^{\dagger}$
\thanks{$^{\dagger}$Xuan Lin conducts this research as an independent researcher {\tt\small \{xuanlin1991\}@gmail.com}}
}

\begin{document}
\maketitle
\thispagestyle{empty}
\pagestyle{empty}

\def\matt#1{\begin{bmatrix}#1\end{bmatrix}}

\begin{abstract}
Hybrid model predictive control (MPC) with both continuous and discrete variables is widely applicable to robotic control tasks, especially those involving contact with the environment. Due to the combinatorial complexity, the solving speed of hybrid MPC can be insufficient for real-time applications. In this paper, we proposed a hybrid MPC solver based on Generalized Benders Decomposition (GBD) with continual learning. The algorithm accumulates cutting planes from the invariant dual space of the subproblems. After a short cold-start phase, the accumulated cuts provide warm-starts for the new problem instances to increase the solving speed. Despite the randomly changing environment that the control is unprepared for, the solving speed maintains. We verified our solver on controlling a cart-pole system with randomly moving soft contact walls and show that the solving speed is 2-3 times faster than the off-the-shelf solver Gurobi.
\end{abstract}
%
%

\section{Introduction} \label{Sec:introduction}
Hybrid model predictive control (MPC) with both continuous and discrete variables is widely applicable to robotic control tasks, especially those involving contact with the environment. However, discontinuous variables are oftentimes computed offline for Hybrid MPC \cite{kuindersma2016optimization, lin2019optimization, hogan2020reactive, zhang2021transition} due to their combinatorial complexities. These include gaits for legged robots and contact sequences for manipulation tasks. Several models with mixed discrete-continuous variables were proposed including mixed-logic dynamic systems (MLDs) \cite{bemporad1999control}, linear complementary models (LCs) \cite{heemels2000linear}, and piece-wise affine systems (PWAs) \cite{sontag1981nonlinear}. Their conditional equivalences were established in \cite{heemels2001equivalence} (for example, LCs are equivalent to MLDs provided that the complementary variables are bounded). Despite several recent works that solve MPC on these systems \cite{aydinoglu2023consensus, cleac2021fast, marcucci2020warm}, the problems addressed in those papers are demonstrate in static environments. In real robotic applications, it is beneficial to further increase the solving speed to reduce the control error since the models are never accurate.

In this paper, we propose a novel hybrid MPC solver based on Generalized Benders decomposition (GBD) \cite{geoffrion1972generalized} to solve problems including MLD constraints under changing environments. Benders decomposition separates the problem into a master problem which solves part of the variables named complicating variables, and a subproblem which solves the rest of the variables. It uses a delayed constraint generation technique that builds up representations of the feasible region and optimal cost function for complicating variables inside the master problem. These representations are constructed as cutting planes based on the dual solutions of the subproblem. The key idea we propose is to accumulate the cuts as more problem instances are solved since the dual feasible set is invariant under the changing environments. The accumulated cuts then provide warm-starts for new problem instances. As a result, the solving speed keeps on increasing. Under the fastest situation, GBD only needs to solve one master problem and one subproblem to get a globally optimal solution. The proposed solver is compared against the recent works on warm-started Branch-Bound solvers and the commercialized solver Gurobi.

We list the contributions below:
\begin{enumerate}
    \item We propose a novel solver based on Generalized Benders Decomposition for hybrid MPCs, where the accumulated cuts provide warm-starts for the new problem instances leading to faster solving speeds despite changing environments. 
    \item We tested our solver on controlling a cart-pole system with randomly moving soft contact walls, a more challenging test than cart-pole with static walls prevail in previous literature \cite{marcucci2020warm, deits2019lvis, aydinoglu2023consensus, cauligi2020learning}. We show that our GBD solver runs faster on average than warm-started Branch and Bound solvers and the off-the-shelf solver Gurobi.
\end{enumerate}


\emph{Notations}
Vectors are bold lowercase; matrices are bold uppercase; sets are script or italicized uppercase.
The real number set is $\mathbb{R}$.
For $\boldsymbol{x}, \boldsymbol{y} \in \mathbb{R}^n$, $\boldsymbol{x} \leq \boldsymbol{y}$ indicates element-wise inequality.
For $\boldsymbol{A} \in \mathbb{R}^{n \times n}$ and $\boldsymbol{B} \in \mathbb{R}^{m \times m}$, $\text{diag}(\boldsymbol{A}, \boldsymbol{B}) \in \mathbb{R}^{(n+m) \times (n+m)}$ denotes the block diagonal matrix with diagonal blocks $\boldsymbol{A}$ and $\boldsymbol{B}$, and zeros otherwise. $\boldsymbol{I}_{n}$ denotes an identity matrix of dimension $n$. The open ball $\textit{B}_{\epsilon}(\boldsymbol{p})$ denotes $\{\boldsymbol{q}: ||\boldsymbol{p} - \boldsymbol{q}|| < \epsilon \}$.

\section{Related Works} \label{Sec:related_work}
\subsection{Mixed-Logic Dynamic Models (MLDs)}
In \cite{bemporad1999control}, the authors proposed mixed-logic dynamics systems as a general modeling tool for control systems incorporating physical laws, logic rules and operating constraints. MLD is rich enough to incorporate dynamic systems such as finite state machines, nonlinear systems that can be written as piece-wise linear form. MLDs have been widely used to model energy storage systems \cite{tobajas2022resilience}, transportation systems \cite{cao2022trajectory}, temperature management systems \cite{klauvco2014building}, to name a few. Recently, MLDs and its equivalent models such as linear complementary models are introduced into the robotics locomotion and manipulation community to model real-time control involving contacts \cite{marcucci2020warm, cleac2021fast, aydinoglu2023consensus, cauligi2021coco}. 

MLDs incorporate states and inputs that can be a mixture of continuous and discrete variables, and quadratic objective functions. Since MLDs incorporate a mixture of discrete and continuous variables, solving it online for fast motion planning and model-predictive control demands an efficient MIQP solver. Several methods have been proposed including explicit MPC \cite{marcucci2017approximate}, Branch-and-Bound \cite{marcucci2020warm}, ADMM \cite{aydinoglu2023consensus}, Lagrange relaxation \cite{geoffrion2010lagrangian}, elastic modes \cite{anitescu2005using}, and generalized Benders decomposition \cite{lazimy1982mixed}. We briefly go over them below.

Explicit MPC solves the problem completely or partially offline, such that the solver only picks out solutions online. \cite{bemporad2002explicit} building polyhedron regions of invariant optimal control function for LQR problems. Explicitly building polytope regions is computationally expensive hence limited to simple problems. \cite{zhu2020fast} stored all solutions and used a K-nearest-neighbor approach to pick out binary solutions. \cite{marcucci2017approximate} explored a partial explicit approach that combines offline solved binary solutions with online convex computation. Related to this work is the library building approach where the problem is solved offline and recorded into a dataset. The solver then picks out data points and uses them as warm-starts. The online data selection approach can be a K-nearest neighbor classifier \cite{lin2022reduce, lin2022learning}, or a learned neural-network \cite{cauligi2021coco}. However, this approach generally has difficulty facing out-of-distribution scenarios from the dataset. Except for \cite{marcucci2020warm}, the works mentioned above only use offline optimal solutions. The infeasible or suboptimal solutions are not used.

Branch and bound is a common approach to solve mixed-integer programs used by off-the-shelf solvers such as Gurobi. This approach first relaxes the integer programming problem into a convex programming on the root node where all binaries are extended to continuous variables between zero and one. It then fixes the binary variables one-by-one, and generate branches from the root to a solution on the leaf node where all binary variables are fixed. Previous studies tried to use information from previous solve to warm-start the new solve online. \cite{hespanhol2019structure} studied propagating the path from root to leaf from the previous iteration to the next one for warm-start, such that a number of parent nodes do not need to be resolved. \cite{marcucci2020warm} further explored propagating complete B\&B tree to warm-start the next problem. Even with the proposed techniques, B\&B can still be slow as it has too many subproblems to keep track of, particularly under noise and model inaccuracies.

Another approach to solve mixed-integer programs is through alternating direction method of multipliers (ADMM). ADMM solves two or multiple problems iteratively until they reach a consensus through a penalty in the objective function. Computer vision and operation research community has used ADMM to solve large scale MIP problems \cite{wu2018ell}. In the robotics community, ADMM has been implemented for solving complementary control problems \cite{aydinoglu2023consensus} at a fast speed. Despite \cite{aydinoglu2023consensus} does not discuss it, ADMM allows for easy warm-start \cite{stellato2020osqp}, using previous solution to accelerate the solving of the next solution. On the other hand, ADMM does not have convergence guarantee for MIP problems unless special assumptions are made such as \cite{wu2018ell}. 


\subsection{Benders Decomposition}
Benders decomposition \cite{benders1962partitioning} can be regarded as Danzig-Wolfe decomposition \cite{dantzig1960decomposition} applied to the dual. Both of them use delayed column or constraint generation techniques. Benders decomposition identifies the complicating variables and defines a subproblem such that those variables are fixed. For this technique to work well, the subproblem should be much easier to solve than the complete problem. For mixed-integer programming, the subproblem is the convex part with complicating variables being the discrete variables \cite{lazimy1982mixed}. As subproblems are solved, cutting planes are added to the master problem to build the feasible set and optimal cost function for the subproblem.  

Benders decomposition was originally proposed to solve linear duals e.g. MILPs. In \cite{geoffrion1972generalized}, the author proposed Generalized Benders decomposition (GBD) that extends the theory to nonlinear duals. Several authors have investigated solving MIQPs using GBD \cite{lazimy1982mixed, watters1967reduction, glover1975improved, mcbride1980implicit}. In \cite{hooker1995logic, hooker2003logic}, the authors propose logic-based Benders decomposition which further generalized the theory to so-called inference dual, which is a logic combination of propositions. This method extends the application of BD to planning and scheduling problems such as satisfiability of 0-1 programming problems whose dual is not a traditional linear or nonlinear programming problem. Using this idea, \cite{codato2006combinatorial} proposed a formulation of combinatorial Benders feasibility cut for MILPs that does not depend on the big-M constant.

As Benders decomposition involves master-subproblem structure, it suits the large-scale distributed problems, or problems with a large number of possible scenarios like stochastic programs \cite{birge2011introduction}. For applications such as distributed control \cite{morocsan2011distributed}, the subproblems can be decoupled into multiple smaller-scale problems and solved in parallel to reduce the computation demand. As pointed out by the review paper \cite{rahmaniani2017benders}, many authors report over 90\% solving time spent on the master problem. Therefore, a number of previous work investigated on how to speed up the master problem, or use its results more efficiently. Examples include local branching heuristics \cite{rei2009accelerating}, heuristic master problem solutions \cite{costa2012accelerating}, generating pareto-optimal cuts \cite{magnanti1981accelerating}, cut initialization \cite{cordeau2001benders}, valid inequalities \cite{rodriguez2021accelerating}, etc. See \cite{rahmaniani2017benders} for a comprehensive review of these methods. \cite{grothey1999note} points out that classic Benders feasibility cuts do not carry objective function value leading to convergence issues. They proposed additional feasibility cuts to resolve this issue. 


GBD can also be used to learn objective functions. This has been applied to dual dynamic programming for MPC over long-term problems \cite{pereira1985stochastic, pereira1991multi}. Previous work \cite{warrington2019generalized, menta2021learning} uses Benders cuts to construct lower bounds for infinitely long objective functions using Bellman operators for both nonlinear and mixed-integer linear systems. Through learning Benders cuts from offline dataset, one avoids hand-tuning terminal cost of objective function. Despite a more optimal objective being learned, the online solving speed of MIP is invariant of objective functions.


\section{Problem Model}
\label{Sec:problem_model}
We develop MPC control laws for Mixed Logic Dynamic (MLD) systems as proposed by \cite{bemporad1999control}:


\begin{subequations}
\begin{align}
    \boldsymbol{x}_{k+1} &= \boldsymbol{E} \boldsymbol{x}_{k} + \boldsymbol{F} \boldsymbol{u}_k + \boldsymbol{G} \boldsymbol{\delta}_k + \boldsymbol{n}_k \\
    \boldsymbol{H}_1 \boldsymbol{x}_{k} &+ \boldsymbol{H}_2 \boldsymbol{u}_k + \boldsymbol{H}_3 \boldsymbol{\delta}_k \leq \boldsymbol{h}(\boldsymbol{\theta}) \label{Eqn:MLD_1b}
\end{align}    
\end{subequations}

At time $k$, $\boldsymbol{x}_k \in \mathbb{R}^{n_{x}}$ is the continuous state. $\boldsymbol{u}_k \in \mathbb{R}^{n_u}$ denotes the continuous input. $\boldsymbol{\delta}_k \in \{0, 1\}^{n_\delta}$ is the binary input. $\boldsymbol{n}_k \in \mathbb{R}^{n_{x}}$ is the disturbance input. Matrices representing system dynamics are $\boldsymbol{E} \in \mathbb{R}^{n_{x} \times n_{x}}$, $\boldsymbol{F} \in \mathbb{R}^{n_{x} \times n_{u}}$, $\boldsymbol{G} \in \mathbb{R}^{n_{x} \times n_{\delta}}$. $\boldsymbol{H}_1 \in \mathbb{R}^{n_c \times n_{x}}$. $\boldsymbol{H}_2 \in \mathbb{R}^{n_c \times n_u}$. $\boldsymbol{H}_3 \in \mathbb{R}^{n_c \times n_{\delta}}$. The submatrices are of appropriate dimensions. The right-hand side of the constraint ~\eqref{Eqn:MLD_1b} is $\boldsymbol{h} \in \mathbb{R}^{n_c}$ where $n_{c}$ is the number of inequality constraints. $\boldsymbol{\theta}$ parameterizes $\boldsymbol{h}$ to represent the changing environments. We assume that matrices $\boldsymbol{E}$, $\boldsymbol{F}$, $\boldsymbol{G}$, $\boldsymbol{H}_1$, $\boldsymbol{H}_2$, $\boldsymbol{H}_3$ are independent of $\boldsymbol{\delta}_k$ and $\boldsymbol{\theta}$. This makes $\boldsymbol{\delta}_k$ and $\boldsymbol{\theta}$ as inputs to the system while the inherent physics of the system is invariant. 

\begin{remark}
The goal of parameter $\boldsymbol{\theta}$ is to represent a sudden change in the environment that the controller is uninformed of and cannot prepare for it down the MPC horizon. Note this is different from the time-varying system investigated by \cite{marcucci2020warm} where the controller is well-informed of the change in advance ($\boldsymbol{\theta}_{t}, t=0,...,T$ is known).
\end{remark}

We formulate a hybrid MPC for this system. The MPC formulation solves an optimization problem to get a sequence of control inputs. However, only the first one is used. It then takes the sensor feedback and resolves the problem. If this could be done fast enough on the hardware, the robot can reject disturbances. The MPC formulation is: 

\begin{alignat*}{2}
    \underset{\boldsymbol{x}_k \in X_k, \ \boldsymbol{u}_k, \ \boldsymbol{\delta}_k}{\text{minimize}}& \ \  \sum_{k=0}^{N-1} \boldsymbol{x}_{k}^{T} \boldsymbol{Q}_k \boldsymbol{x}_k + \boldsymbol{u}_{k}^{T} \boldsymbol{R}_k \boldsymbol{u}_k + \boldsymbol{x}_{N}^{T} \boldsymbol{Q}_N \boldsymbol{x}_N\\
    \text{s.t.} \ \ &\boldsymbol{x}_0 = \boldsymbol{x}_{ini} \\
    & \boldsymbol{x}_{k+1} = \boldsymbol{E} \boldsymbol{x}_k + \boldsymbol{F} \boldsymbol{u}_k + \boldsymbol{G} \boldsymbol{\delta}_k \\
    & \boldsymbol{H}_{1} \boldsymbol{x}_{k} + \boldsymbol{H}_{2} \boldsymbol{u}_k + \boldsymbol{H}_3 \boldsymbol{\delta}_k \leq \boldsymbol{h}(\boldsymbol{\theta}) \\
    & \boldsymbol{\delta}_k \in \{0, 1\}^{n_{\delta}}, \ k=0,...,N-1
\end{alignat*}
\label{Eqn:MLD}

The matrices $\boldsymbol{Q}_k$ and $\boldsymbol{Q}_{N}$ are positive definite matrices. $X_k$ is the domain of $\boldsymbol{x}_k$. The system is written into a more compact form:

\begin{equation}
\begin{aligned}
   \underset{\boldsymbol{x} \in X, \ \boldsymbol{\delta}}{\text{minimize}} \ \ & \boldsymbol{x}^{T} \boldsymbol{Q} \boldsymbol{x} \\
    \text{s.t.} \ \ & \boldsymbol{A} \boldsymbol{x} = \boldsymbol{b}(\boldsymbol{x}_{ini}, \boldsymbol{\delta})\\
    & \boldsymbol{C} \boldsymbol{x} \leq \boldsymbol{d}( \boldsymbol{\theta}, \boldsymbol{\delta}) \\
    & \boldsymbol{\delta}_k \in \{0, 1\}^{n_{\delta}}
\end{aligned}
\label{Eqn:MLD_compact}
\end{equation}

Let $n_{xu} = n_x + n_u$. The matrices and vectors have structures:

\begin{subequations}
\begin{align}
    \boldsymbol{x} &= \matt{\boldsymbol{x}^T_0 & \boldsymbol{u}^T_0 & \cdots & \boldsymbol{x}^T_{N-1} & \boldsymbol{u}^T_{N-1} & \boldsymbol{x}^T_{N}}^T \in \mathbb{R}^{Nn_{xu}+n_x} \\
\boldsymbol{\delta} &= \matt{\boldsymbol{\delta}^T_0 & \cdots & \boldsymbol{\delta}^T_{N}}^T \in \mathbb{R}^{(N+1)n_{\delta}}
\end{align}
\end{subequations}

Hence domain of $\boldsymbol{x}$ is $X = X_k \times \mathbb{R}^{n_u} \times \cdots \times \mathbb{R}^{n_u} \times X_k$. 

\begin{equation}
\begin{aligned}
    \boldsymbol{A} = &\begin{bmatrix}
    \boldsymbol{I}_{n_{x}} \ \ \boldsymbol{0} \\
    -\boldsymbol{E} \  -\boldsymbol{F}   & \boldsymbol{I}_{n_{x}} \ \ \boldsymbol{0} \\
                                  & -\boldsymbol{E} \ -\boldsymbol{F} & \ddots \\         
                                  &                             & \ddots   & \boldsymbol{I}_{n_{x}} \ \ \boldsymbol{0}      \\
                                  &                             &          & -\boldsymbol{E} \ -\boldsymbol{F}   & \boldsymbol{I}_{n_{x}}
    \end{bmatrix} \\
    & \in \mathbb{R}^{(N+1)n_x \times (Nn_{xu}+n_x)}
\end{aligned}
\end{equation}

\begin{equation}
\begin{aligned}
    \boldsymbol{b}(\boldsymbol{x}_{ini}, \boldsymbol{\delta}) &= \matt{\boldsymbol{x}_{ini}^T & (\boldsymbol{G}\boldsymbol{\delta}_0)^T & \cdots & (\boldsymbol{G}\boldsymbol{\delta}_{N-1})^T}^T \\ &\in \mathbb{R}^{(N+1)n_x}
\end{aligned}
\end{equation}

\begin{equation}
\begin{aligned}
    \boldsymbol{C} = &\begin{bmatrix}
    \boldsymbol{H}_1 \ \boldsymbol{H}_2 \\
                     & \boldsymbol{H}_1 \ \boldsymbol{H}_2 \\
                     &                  & \ddots \\         
                     &                  &        & \boldsymbol{H}_1 \ \boldsymbol{H}_2 & \boldsymbol{0}
    \end{bmatrix} \\
    & \in \mathbb{R}^{Nn_{c} \times (Nn_{xu}+n_x)}
\end{aligned}
\end{equation}

\begin{equation}
\begin{aligned}
    \boldsymbol{d}(\boldsymbol{\theta}, \boldsymbol{\delta}) &=
    \matt{ (\boldsymbol{h}(\boldsymbol{\theta}) - \boldsymbol{H}_3\boldsymbol{\delta}_0)^T & \cdots & (\boldsymbol{h}(\boldsymbol{\theta}) - \boldsymbol{H}_3\boldsymbol{\delta}_{N-1})^T}^T \\
    & \in \mathbb{R}^{Nn_{c}}
\end{aligned}
\end{equation}

\begin{equation}
    \boldsymbol{Q} = \text{diag}(\boldsymbol{Q}_k, \boldsymbol{R}_k) \in \mathbb{R}^{(Nn_{xu}+n_x) \times (Nn_{xu}+n_x)}  
\end{equation}

Problem ~\eqref{Eqn:MLD_compact} is an MIQP and can be solved through an off-the-shelf mixed-integer convex programming solver based on Branch and Bound, such as Gurobi. However, Branch and Bound algorithms keep track of a large number of subproblems that relax the binary constraints in different ways. Despite the MPC warm-start scheme such as shifting contact sequence can be used \cite{marcucci2020warm}, many subproblems still need to be solved for a new problem instance. For applications that require extremely fast solving speed, this can be insufficient. In this paper, we propose to use generalized Benders decomposition to solve the problem several times faster than Gurobi.

\section{Benders decomposition formulation} 
\label{Sec:Benders_formulation}
In this section, we apply Benders decomposition to our hybrid MPC problem ~\eqref{Eqn:MLD}. Benders decomposition deals with the problem of the following form:

\begin{equation}
\begin{aligned}
    \underset{\boldsymbol{x}, \boldsymbol{y}}{\text{minimize}} \ \ f(\boldsymbol{x}, \boldsymbol{y}) \\
    \text{s.t.} \ \ \boldsymbol{G}(\boldsymbol{x}, \boldsymbol{y}) \leq 0 \\
    \boldsymbol{x} \in X, \boldsymbol{y} \in Y
\end{aligned} 
\label{Eqn:original}
\end{equation}

where $\boldsymbol{y}$ is a vector of complicating variables. If $\boldsymbol{y}$ is fixed, the optimization problem is much easier to solve. Benders decomposition partitions the problem into a master problem by projecting onto the $\boldsymbol{y}$ space:

\begin{equation}
\begin{aligned}
    \underset{\boldsymbol{y}}{\text{minimize}} \ \ v(\boldsymbol{y}) \\
    \text{s.t.} \ \ \boldsymbol{y} \in Y \cap V
\end{aligned} 
\label{Eqn:master_prob}
\end{equation}

The function $v(\boldsymbol{y})$ is defined to provide the best objective function with fixed complicating variable $\boldsymbol{y}$:

\begin{equation}
\begin{aligned}
    v(\boldsymbol{y}) = \underset{\boldsymbol{x}}{\text{infimum}} \ \ f(\boldsymbol{x}, \boldsymbol{y}) \\
                    \text{s.t.} \ \ \boldsymbol{G}(\boldsymbol{x}, \boldsymbol{y}) \leq 0 \\
                                    \boldsymbol{x} \in X
\end{aligned} 
\label{Eqn:sub_prob}
\end{equation}

$V$ contains all $\boldsymbol{y}$'s such that problem ~\eqref{Eqn:sub_prob} is feasible:

\begin{equation}
    V = \{\boldsymbol{y}: \boldsymbol{G}(\boldsymbol{x}, \boldsymbol{y}) \leq 0, \ \ \exists \boldsymbol{x} \in X\}
\end{equation}

For our hybrid MPC, we define the complicating variable $\boldsymbol{y}$ as the binary variable $\boldsymbol{\delta}$, the initial condition $\boldsymbol{x}_{ini}$, and the parameter $\boldsymbol{\theta}$. The subproblem is:

\begin{equation}
\begin{aligned}
   v(\boldsymbol{x}_{ini}, \boldsymbol{\theta}, \boldsymbol{\delta}) =  \underset{\boldsymbol{x} \in X}{\text{minimize}} \ \ & \boldsymbol{x}^{T} \boldsymbol{Q} \boldsymbol{x} \\
    \text{s.t.} \ \ & \boldsymbol{A} \boldsymbol{x} = \boldsymbol{b}(\boldsymbol{x}_{ini}, \boldsymbol{\delta})\\
    & \boldsymbol{C} \boldsymbol{x} \leq \boldsymbol{d}(\boldsymbol{\theta}, \boldsymbol{\delta})
\end{aligned}
\label{Eqn:subproblem_MLD}
\end{equation}

Given fixed $\boldsymbol{x}_{ini}$, $\boldsymbol{\theta}$, $\boldsymbol{\delta}$, ~\eqref{Eqn:subproblem_MLD} is a quadratic programming and can be solved through off-the-shelf QP solvers. The master problem is:

\begin{equation}
\begin{aligned}
    \underset{\boldsymbol{\delta}}{\text{minimize}} \ \ &v(\boldsymbol{x}_{ini}, \boldsymbol{\theta}, \boldsymbol{\delta}) \\
    \text{s.t.} \ \ &\boldsymbol{\delta}_k \in \{0, 1\} \\
    & \boldsymbol{\delta} \in V \coloneqq \{\boldsymbol{\delta}: \boldsymbol{A} \boldsymbol{x} = \boldsymbol{b}(\boldsymbol{x}_{ini}, \boldsymbol{\delta})\\
    & \boldsymbol{C} \boldsymbol{x} \leq \boldsymbol{d}(\boldsymbol{\theta}, \boldsymbol{\delta}), \exists 
 \boldsymbol{x} \in X\} \\
\end{aligned} 
\label{Eqn:masterproblem_MLD}
\end{equation}

The essential issue with solving ~\eqref{Eqn:masterproblem_MLD} is that function $v(\boldsymbol{x}_{ini}, \boldsymbol{\theta}, \boldsymbol{\delta})$ and set $V$ are only implicitly known through their definitions. Benders decomposition is a process that iteratively solves problem ~\eqref{Eqn:master_prob} and ~\eqref{Eqn:sub_prob} to build approximations of $v$ and $V$ in the problem ~\eqref{Eqn:master_prob}.

We will constantly work with the dual of problem ~\eqref{Eqn:subproblem_MLD}, given the advantage that the dual is invariant with respect to the complicating variables. We derive the dual for reference. Recall the definition of Lagrangian for the subproblem ~\eqref{Eqn:subproblem_MLD}:

\begin{equation}
\begin{aligned}
    \mathcal{L}(\boldsymbol{x}, \boldsymbol\nu, \boldsymbol\lambda; \boldsymbol{x}_{ini}, \boldsymbol{\theta}, \boldsymbol{\delta}) = f_{obj}(\boldsymbol{x}) &+ \boldsymbol\nu^T (\boldsymbol{A} \boldsymbol{x} - \boldsymbol{b}(\boldsymbol{x}_{ini}, \boldsymbol{\delta})) \\
    &+ \boldsymbol\lambda^T(\boldsymbol{C} \boldsymbol{x} - \boldsymbol{d}(\boldsymbol{\theta}, \boldsymbol{\delta})) 
\end{aligned}
\label{Eqn:Lagrangian}
\end{equation}

where $\boldsymbol\nu \in \mathbb{R}^{(N+1)n_{x}}$, $\boldsymbol\lambda \in \mathbb{R}^{Nn_{c}}$ are the dual variables associated with $\boldsymbol{A} \boldsymbol{x} = \boldsymbol{b}(\boldsymbol{x}_{ini}, \boldsymbol{\delta})$, $\boldsymbol{C} \boldsymbol{x} \leq \boldsymbol{d}( \boldsymbol{\theta}, \boldsymbol{\delta})$, respectively. The Lagrange dual function $g$ is:

\begin{equation}
    g(\boldsymbol\nu, \boldsymbol\lambda; \boldsymbol{x}_{ini}, \boldsymbol{\theta}, \boldsymbol{\delta}) = \underset{\boldsymbol{x} \in X}{\text{minimize}} \ \ \mathcal{L}(\boldsymbol{x}, \boldsymbol\nu, \boldsymbol\lambda; \boldsymbol{x}_{ini}, \boldsymbol{\theta}, \boldsymbol{\delta})
\end{equation}

where $f_{obj} = \boldsymbol{x}^{T} \boldsymbol{Q} \boldsymbol{x}$. 
Let $\boldsymbol{x}^0$ be the unconstrained minimizer of $\mathcal{L}$ (in general, $\boldsymbol{x}^0$ different from the opitmal primal solution $\boldsymbol{x}^*$). By taking derivative, we have:

\begin{equation}
    \boldsymbol{x}^0 = -\frac{1}{2} \boldsymbol{Q}^{-1} (\boldsymbol{A}^T\boldsymbol{\nu} + \boldsymbol{C}^T \boldsymbol{\lambda})
\label{Eqn:unconstrained_optimizer}
\end{equation}

Hence the Lagrange dual problem is:

\begin{equation}
\begin{aligned}
     \underset{\boldsymbol{\nu}, \ \boldsymbol{\lambda}}{\text{maximize}} & \ \ -\frac{1}{4}||\boldsymbol{A}^T\boldsymbol{\nu} + \boldsymbol{C}^T \boldsymbol{\lambda}||^2_{\boldsymbol{Q}^{-1}} \\
     &\ \ - \boldsymbol{b}(\boldsymbol{x}_{ini}, \boldsymbol{\delta})^T \boldsymbol{\nu} - \boldsymbol{d}(\boldsymbol{\theta}, \boldsymbol{\delta})^T \boldsymbol{\lambda} \\
     \text{s.t.} & \ \ \boldsymbol{\lambda} \geq \boldsymbol{0}
\end{aligned}
\label{Eqn:feasible_dual}
\end{equation}

As the feasibility of ~\eqref{Eqn:subproblem_MLD} is independent of the objective function, ~\eqref{Eqn:subproblem_MLD} is feasible if and only if the following problem is feasible: 

\begin{equation}
\begin{aligned}
   \underset{\boldsymbol{x} \in X}{\text{minimize}} \ \ & \boldsymbol{0} \\
    \text{s.t.} \ \ & \boldsymbol{A} \boldsymbol{x} = \boldsymbol{b}(\boldsymbol{x}_{ini}, \boldsymbol{\delta})\\
    & \boldsymbol{C} \boldsymbol{x} \leq \boldsymbol{d}(\boldsymbol{\theta}, \boldsymbol{\delta})
\end{aligned}
\label{Eqn:infeasible_MLD}
\end{equation}

Problem ~\eqref{Eqn:infeasible_MLD} has the dual:
\begin{equation}
\begin{aligned}
    \underset{\boldsymbol\nu, \ \boldsymbol\lambda}{\text{maximize}} &\ \ - \boldsymbol{b}(\boldsymbol{x}_{ini}, \boldsymbol{\delta})^T \boldsymbol{\nu} - \boldsymbol{d}(\boldsymbol{\theta}, \boldsymbol{\delta})^T \boldsymbol{\lambda} \\
    \text{s.t.} \ \ &\boldsymbol{A}^T\boldsymbol{\nu} + \boldsymbol{C}^T \boldsymbol{\lambda} = \boldsymbol{0} \\
    &\boldsymbol\lambda \geq \boldsymbol{0}
\end{aligned}
\label{Eqn:Infeasible_dual}
\end{equation}

\subsection{Feasibility cuts}
If at iteration $p$, the subproblem is infeasible under the given $\boldsymbol{\delta}_p$, this $\boldsymbol{\delta}_p$ needs to be removed from the master problem. This can be achieved by adding a cutting plane. Since the problem ~\eqref{Eqn:MLD} is linearly constrained, the Farkas certificates can be used to add feasibility cuts. They can be discovered by solving ~\eqref{Eqn:infeasible_MLD} with a dual simplex solver (\cite{bertsimas1997introduction}, Chapter 6.5). The theorem of alternatives for ~\eqref{Eqn:infeasible_MLD} is:

\begin{lemma} 
\label{Lem:lemma1}
Given $\boldsymbol{A} \in \mathbb{R}^{l \times n}$, $\boldsymbol{b} \in \mathbb{R}^{l}$, $\boldsymbol{C} \in \mathbb{R}^{m \times n}$, $\boldsymbol{d} \in \mathbb{R}^{m}$, exactly one of the following statements is true:
\begin{enumerate}
    \item There exists an $\boldsymbol{x} \in \mathbb{R}^{n}$ that satisfies $\boldsymbol{A}\boldsymbol{x} = \boldsymbol{b}, \boldsymbol{C}\boldsymbol{x} \leq \boldsymbol{d}$.
    \item There exist $\boldsymbol{y} \in \mathbb{R}^{l}$, $\boldsymbol{z} \in \mathbb{R}^{m}$ that satisfy $\boldsymbol{z} \geq \boldsymbol{0}$, $\boldsymbol{A}^T \boldsymbol{y} + \boldsymbol{C}^T \boldsymbol{z} = \boldsymbol{0}$, $\boldsymbol{b}^T \boldsymbol{y} + \boldsymbol{d}^T \boldsymbol{z} < 0$.
\end{enumerate}
\end{lemma}

\begin{proof}
    See Appendix \ref{Appendix_pf_lemma1}.
\end{proof}


If ~\eqref{Eqn:infeasible_MLD} is infeasible for $\boldsymbol{\delta}_p$. Then we can add a cutting plane to the master problem to remove a set of $\boldsymbol{\delta}$'s including $\boldsymbol{\delta}_p$. Farkas lemma guarantees the existence of $\tilde{\boldsymbol\nu}_{p} \in \mathbb{R}^{(N+1)n_{x}}$, $\tilde{\boldsymbol\lambda}_{p} \in \mathbb{R}^{Nn_{c}}$ such that:

\begin{equation}
\begin{aligned}
    &\tilde{\boldsymbol\lambda}_{p} \geq \boldsymbol{0} \\ &\boldsymbol{A}^{T} \tilde{\boldsymbol\nu}_{p} + \boldsymbol{C}^{T} \tilde{\boldsymbol\lambda}_{p} = \boldsymbol{0} \\
    &\boldsymbol{b}(\boldsymbol{x}_{ini}, \boldsymbol{\delta}_p)^{T} \tilde{\boldsymbol\nu}_{p} + \boldsymbol{d}(\boldsymbol{\theta}, \boldsymbol{\delta}_p)^{T} \tilde{\boldsymbol\lambda}_{p} < 0
\end{aligned}
\label{Eqn:Farkas_proof}
\end{equation}

To prevent the master problem from giving this $\boldsymbol{\delta}_p$, a constraint to defeat the Farkas infeasible proof is added to the master problem:

\begin{equation}
    \boldsymbol{b}(\boldsymbol{x}_{ini}, \boldsymbol{\delta})^{T} \tilde{\boldsymbol\nu}_{p} + \boldsymbol{d}(\boldsymbol{\theta}, \boldsymbol{\delta})^{T} \tilde{\boldsymbol\lambda}_{p} \geq 0
\label{Eqn:feasible_cutting_plane}
\end{equation}

We note that this cutting plane will not remove any feasible $\boldsymbol{\delta}$ from the subproblem. We state this as a lemma.

\begin{lemma}
For given $\boldsymbol{x}_{ini}$ and $\boldsymbol{\theta}$, any $\boldsymbol{\delta}$ that contradicts ~\eqref{Eqn:feasible_cutting_plane} proves infeasibility for ~\eqref{Eqn:infeasible_MLD}.
\label{Lemma2}
\end{lemma}

\begin{proof} 
As $\tilde{\boldsymbol\nu}_{p}$, $\tilde{\boldsymbol\lambda}_{p}$ discovered by the dual simplex solver satisfy the first two conditions of ~\eqref{Eqn:Farkas_proof}, they are feasible for the dual problem ~\eqref{Eqn:Infeasible_dual}. Let $a \in \mathbb{R}^{+}$ be an arbitrary positive value, $(a\tilde{\boldsymbol\nu}_{p}, a\tilde{\boldsymbol\lambda}_{p})$ are also feasible for ~\eqref{Eqn:Infeasible_dual}. Let $\boldsymbol{\delta}$ be any value that contradicts ~\eqref{Eqn:feasible_cutting_plane}, we have $-a\tilde{\boldsymbol\nu}_{p}^{T}\boldsymbol{b}(\boldsymbol{x}_{ini}, \boldsymbol{\delta}) - a\tilde{\boldsymbol\lambda}_{p}^{T}\boldsymbol{d}(\boldsymbol{\theta}, \boldsymbol{\delta}) \rightarrow +\infty$ as $a \rightarrow +\infty$, hence the dual problem is unbounded which proves that the primal problem ~\eqref{Eqn:infeasible_MLD} is infeasible (from Corollary 4.1 of \cite{bertsimas1997introduction}). 
\end{proof}

Since hybrid MPC needs to be solved fast online, it is important to maximize the usage of computations so the number of iteration to find a feasible solution is reduced. Many previous works added one feasibility cut each iteration. 
Some previous works \cite{ magnanti1981accelerating, wu2010accelerating, beheshti2019accelerating} propose adding multiple cutting planes each iteration, or re-formulate the problem such that stronger cuts can be generated. However, the subproblem structure has not been explored by those papers. We propose an innovative technique to add multiple feasibility cuts to the master problem via subproblem recursive structure. The online computation time prevents us to solve any additional optimization problems (even convex ones), but those cutting planes can be retrieved without any additional computation given the planes we already have.

Define $\tilde{\boldsymbol\nu}_p^m$, $\tilde{\boldsymbol\lambda}_p^m$, $m=1, ..., N-1$ such that:

\begin{equation}
\begin{aligned}
    \tilde{\boldsymbol\nu}_{p,k}^{m} &= \begin{cases}
                                     \tilde{\boldsymbol\nu}_{p,k+m} \ \ &\forall k+m \leq N \\
                                     \boldsymbol{0} \ \ &\forall k+m > N \\ 
                                     \end{cases} \\
    \tilde{\boldsymbol\lambda}_{p,k}^{m} &= \begin{cases}
                                     \tilde{\boldsymbol\lambda}_{p,k+m} \ \ &\forall k+m \leq N-1 \\
                                     \boldsymbol{0} \ \ &\forall k+m > N-1 \\  
                                     \end{cases} \\
\end{aligned}
\label{Eqn:shifted_cuts}
\end{equation}

For each $m$, we add an additional cutting planes:

\begin{equation}
    \boldsymbol{b}(\boldsymbol{x}_{ini}, \boldsymbol{\delta})^{T} \tilde{\boldsymbol\nu}_{p}^{m} + \boldsymbol{d}(\boldsymbol{\theta}, \boldsymbol{\delta})^{T} \tilde{\boldsymbol\lambda}_{p}^{m} \geq 0, \ \ m=1, ..., N-1
\label{Eqn:feasible_cutting_plane_shifted}
\end{equation}

The addition of cuts ~\eqref{Eqn:feasible_cutting_plane_shifted} works in two ways. First, the solutions that contradicts ~\eqref{Eqn:feasible_cutting_plane_shifted} has good optimality as predicted by the optimality cuts, hence may be selected by the master problem as the next trial solution. Therefore, addition of ~\eqref{Eqn:feasible_cutting_plane_shifted} eliminates those trials and accelerates the master problem to find a feasible solution, especially in the cold start stage when the master problem is almost empty. Second, cuts ~\eqref{Eqn:feasible_cutting_plane_shifted} predicts the future infeasible cased by shifting the current infeasible cases into the future, such that the future solves do not need to re-discover them. Similar to $\tilde{\boldsymbol\nu}_{p}$ and $\tilde{\boldsymbol\lambda}_{p}$, we present:

\begin{corollary}
Any $\boldsymbol{\delta}$ that contradicts ~\eqref{Eqn:feasible_cutting_plane_shifted} proves infeasibility for ~\eqref{Eqn:infeasible_MLD} with given $\boldsymbol{x}_{ini}$ and $\boldsymbol{\theta}$.
\end{corollary}

\begin{proof}
We can verify that $\tilde{\boldsymbol\nu}_{p}^{m}$, $\tilde{\boldsymbol\lambda}_{p}^{m}$ are dual feasible for any $m$ if $\tilde{\boldsymbol\nu}_{p}$, $\tilde{\boldsymbol\lambda}_{p}$ are dual feasible by simply plugging them into the dual feasibility constraints. This is a simple extension of Lemma \ref{Lemma2} .
\end{proof}

For our problem, $\boldsymbol{b}(\boldsymbol{x}_{ini}, \boldsymbol{\delta})$, $\boldsymbol{d}(\boldsymbol{\theta}, \boldsymbol{\delta})$ depend on $\boldsymbol{\delta}$ linearly, it is interesting to realize that from an infeasible subproblem with one $\boldsymbol{\delta}$, we construct a plane that may remove a set of infeasible $\boldsymbol{\delta}$'s. This contributes to the efficacy of Benders decomposition as it takes usage of infeasible samples which are usually thrown away by the methods that learn binary solutions offline \cite{ marcucci2017approximate, cauligi2021coco, lin2022learning}.

\subsection{Optimality cuts}

If at iteration $q$, the sub-problem is solved to optimal under given $\boldsymbol{\delta}_q$, we want to add a cutting plane as a lower bound that approaches $v(\boldsymbol{x}_{ini}, \boldsymbol{\theta}, \boldsymbol{\delta})$ from below. This can be realized through duality theory. For any $\boldsymbol\nu$ and $\boldsymbol\lambda \geq \boldsymbol{0}$, $g(\boldsymbol\nu, \boldsymbol\lambda; \boldsymbol{x}_{ini}, \boldsymbol{\theta}, \boldsymbol{\delta}_q) \leq v(\boldsymbol{x}_{ini}, \boldsymbol{\theta}, \boldsymbol{\delta}_q)$. Since the subproblem is convex and we assume there exists a stricly feasible solution (Slater's), strong duality is achieved and the best lower bound is tight:

\begin{equation}
    v(\boldsymbol{x}_{ini}, \boldsymbol{\theta}, \boldsymbol{\delta}_q) = \underset{\boldsymbol\nu, \boldsymbol\lambda \geq \boldsymbol{0}}{\text{maximize}} \ \ g(\boldsymbol\nu, \boldsymbol\lambda; \boldsymbol{x}_{ini}, \boldsymbol{\theta}, \boldsymbol{\delta}_q)
\end{equation}

Therefore, we add the best lower bound as a cutting plane to the master problem. Let $\boldsymbol{x}_q^0$ be the unconstrained minimizer of $\mathcal{L}$ at iteration $q$, and $\boldsymbol\nu_{q}^*, \boldsymbol\lambda_{q}^*$ be the maximizer of $g(\boldsymbol\nu, \boldsymbol\lambda; \boldsymbol{x}_{ini}, \boldsymbol{\theta}, \boldsymbol{\delta}_q)$, the cutting plane takes the form:

\begin{equation}
\begin{aligned}
    v(\boldsymbol{x}_{ini}, \boldsymbol{\theta}, \boldsymbol{\delta}) &\geq \mathcal{L}(\boldsymbol{x}_q^0, \boldsymbol\nu_{q}^*, \boldsymbol\lambda_{q}^*; \boldsymbol{x}_{ini}, \boldsymbol{\theta}, \boldsymbol{\delta}) \\
    & =  f_{obj}(\boldsymbol{x}_q^0) + \boldsymbol\nu_{q}^{*T} \boldsymbol{A} \boldsymbol{x}_q^0 + \boldsymbol\lambda_{q}^{*T} \boldsymbol{C} \boldsymbol{x}_q^0\\
    & - \boldsymbol\nu_{q}^{*T} \boldsymbol{b}(\boldsymbol{x}_{ini}, \boldsymbol{\delta}) - \boldsymbol\lambda_{q}^{*T} \boldsymbol{d}(\boldsymbol{\theta}, \boldsymbol{\delta})
    \triangleq  \mathcal{L}^*(\boldsymbol{x}_{ini}, \boldsymbol{\theta}, \boldsymbol{\delta})
\end{aligned}
\label{Eqn:optimal_cut_MLD}
\end{equation}

Note that $\boldsymbol\nu_{q}^*, \boldsymbol\lambda_{q}^*$ depend on $\boldsymbol{\delta}_q, \boldsymbol{x}_{ini}$. We make one key observation:

\begin{proposition}
$\boldsymbol{x}_q^0$ depends on $\boldsymbol\nu_{q}^*, \boldsymbol\lambda_{q}^*$ but does not have explicit dependency on $\boldsymbol{\delta}_q, \boldsymbol{x}_{ini}$, $\boldsymbol{\theta}$.
\label{Assumption_1}
\end{proposition}
\begin{proof}
    This is true given ~\eqref{Eqn:unconstrained_optimizer}.
\end{proof}

With Proposition \ref{Assumption_1}, when $\boldsymbol{\delta}_q, \boldsymbol{x}_{ini}$, $\boldsymbol{\theta}$ change, $\boldsymbol{x}_q^0$ is still the unconstrained minimizer of $\mathcal{L}$ as long as we do not swap $\boldsymbol\nu_{q}^*, \boldsymbol\lambda_{q}^*$. However, $\boldsymbol\nu_{q}^*, \boldsymbol\lambda_{q}^*$ are no longer the maximizer of $g$. Hence, $\mathcal{L}^*(\boldsymbol{x}_{ini}, \boldsymbol{\theta}, \boldsymbol{\delta})$ only provides a loose lower bound for $\boldsymbol{\delta}$ other than the current $\boldsymbol{\delta}_q$ used to generate this optimality cut. The subscript $q$ is dropped in ~\eqref{Eqn:optimal_cut_MLD} indicating that the inequality is valid for general $\boldsymbol{\delta}$. As $\boldsymbol{x}_{ini}$ and $\boldsymbol{\theta}$ take the same position as $\boldsymbol{\delta}_q$ in $\mathcal{L}$, the same argument applies when $\boldsymbol{x}_{ini}$ and $\boldsymbol{\theta}$ are updated. This will be used to construct warm-starts for hybrid MPC.

If the solver used for the subproblem does not return the unconstrained optimizer $\boldsymbol{x}_q^0$, we can leverage on strong duality to avoid computing $\boldsymbol{x}_q^0$. Since $v(\boldsymbol{x}_{ini}, \boldsymbol{\theta}, \boldsymbol{\delta}_q) = f_{obj}(\boldsymbol{x}_q^*) \triangleq f_{obj,q}^* = \mathcal{L}^* ( \boldsymbol{x}_{ini}, \boldsymbol{\theta}, \boldsymbol{\delta}_q)$, the cutting plane takes the form:

\begin{equation}
\begin{aligned}
    v(\boldsymbol{x}_{ini}, \boldsymbol{\theta}, \boldsymbol{\delta}) \geq f_{obj,q}^* &+ \boldsymbol\nu_{q}^{*T} (\boldsymbol{b}(\boldsymbol{x}_{ini}, \boldsymbol{\delta}_q) - \boldsymbol{b}(\boldsymbol{x}_{ini}, \boldsymbol{\delta})) \\
    &+ \boldsymbol\lambda_{q}^{*T}(\boldsymbol{d}(\boldsymbol{\theta}, \boldsymbol{\delta}_q)-\boldsymbol{d}(\boldsymbol{\theta}, \boldsymbol{\delta}))
\end{aligned}
\label{Eqn:optimal_cut_simple_MLD}
\end{equation}

\subsection{The Benders master problem}
The final form of the master problem ~\eqref{Eqn:masterproblem_MLD} is:

\begin{equation}
\begin{aligned}
    &\underset{\boldsymbol{\delta}}{\text{minimize}} \ \ z_0 \\
    \text{s.t.} \ \ &\boldsymbol{\delta}_k \in \{0, 1\} \\
    & for \ p=1, ..., \text{current \# of infeasible subproblem:} \\
    &\boldsymbol{b}(\boldsymbol{x}_{ini}, \boldsymbol{\delta})^{T} \tilde{\boldsymbol\nu}_{p} + \boldsymbol{d}(\boldsymbol{\theta}, \boldsymbol{\delta})^{T} \tilde{\boldsymbol\lambda}_{p} \geq 0 \\
    & \boldsymbol{b}(\boldsymbol{x}_{ini}, \boldsymbol{\delta})^{T} \tilde{\boldsymbol\nu}_{p}^{m} + \boldsymbol{d}(\boldsymbol{\theta}, \boldsymbol{\delta})^{T} \tilde{\boldsymbol\lambda}_{p}^{m} \geq 0, \ \ m=1, ..., N-1 \\
    & for \ q=1, ..., \text{current \# of optimal subproblem:} \\
    & z_0 \geq f_{obj,q}^* + \boldsymbol\nu_{q}^{*T} (\boldsymbol{b}(\boldsymbol{x}_{ini}, \boldsymbol{\delta}_q) - \boldsymbol{b}(\boldsymbol{x}_{ini}, \boldsymbol{\delta})) + \\
    & \ \ \ \ \ \ \ \boldsymbol\lambda_{q}^{*T} (\boldsymbol{d}(\boldsymbol{\theta}, \boldsymbol{\delta}_q)-\boldsymbol{d}(\boldsymbol{\theta}, \boldsymbol{\delta})) \\
\end{aligned} 
\label{Eqn:masterproblem_MLD_final}
\end{equation}

We define the epigraph variable $z_0$ and make $z_0 \geq v(\boldsymbol{x}_{ini}, \boldsymbol{\theta}, \boldsymbol{\delta}_q) \ \forall q$ to find the smallest value of optimality cuts corresponding to binary mode $\boldsymbol{\delta}$. For our MPC problem with relatively small $N$ ($10 \sim 30$), this problem is a small-scale MIP problem that contains $2N$ binary variables and one continuous variable. As the algorithm proceeds, constraints will be added to the master problem. For our test, the number of feasibility cuts and optimality cuts has a scale of hundreds or thousands. The master problem is solved within the presolve stage by an off-the-shelf MIP solver.

\subsection{Upper bound and lower bound of the cost}
Benders decomposition is an iterative procedure that adds feasibility cuts and optimality cuts to improve the approximation of $v$ and $V$ in the master problem. Since the optimality cuts are lower bounds of the actual cost $v$ according to ~\eqref{Eqn:optimal_cut_MLD}, the master problem will "underestimate the difficulty" of the subproblem when few cuts are generated. This means it will propose $\boldsymbol{\delta}$ that is oftentimes infeasible, or with an optimal cost cannot be reached by the subproblem. We define the master problem cost $z_{0,i}$ at the latest iteration $i$ as the lower bound of the global optimal cost. As more cuts are added, the lower bound increases to approach the actual cost. Since at each iteration $i$, the MIP solver finds the global optimal value of the master problem, the lower bound does not decrease with respect to $i$ (if at next iteration $i+1$, $z_{0,i+1} < z_{0,i}$, $z_{0,i+1}$ should be achievable at iteration $i$ since the problem has one additional constraint at iteration $i+1$ than $i$, contradicting the fact that MIP solvers find the global optimal solution).

On the other hand, the best subproblem cost till the current iteration provides a upper bound of the global optimal cost. Since the accumulated cuts in the master problem underestimate the actual subproblem cost, the cost of the iteration $i+1$, $f^*_{obj,i+1}$, may not decrease comparing to the cost $f^*_{obj,i}$ at iteration $i$, if the estimated cost is far from the actual one. The upper bound is introduced to keep track of the best cost. 

As the algorithm proceeds, the lower bound and upper bound will approach each other. Eventually, they converge to each other and the algorithm terminates (a proof of convergence can be found in \cite{geoffrion1972generalized}). We introduce the $MIPgap$ for the termination condition. For fair comparison, we use the same definition as Gurobi \cite{MIPGap}:

\begin{equation}
    g_a = |z_P-z_D|/|z_P|
\end{equation}

Where $z_P$ is the primal bound, or upper bound of the global optimal cost, and $z_D$ is the dual bound, or the lower bound of the cost. When $g_a$ is smaller than a predefined threshold $G_a$, the inner loop algorithm terminates with an optimal $\boldsymbol{\delta}^*$ and control signal $\boldsymbol{u}^*$. We present the Benders MPC algorithm in Algorithm \ref{Initial_algorithm}.

\begin{algorithm}
\caption{Benders MPC}
\label{Initial_algorithm}
\KwIn{$G_a$}
\textit{Initialization} $LB \coloneqq -\infty$, $UB \coloneqq \infty$, iteration $i \coloneqq 0$\\
\While{$|UB-LB|/|UB| \geq G_a$}{
Solve master problem ~\eqref{Eqn:masterproblem_MLD_final} to get $\boldsymbol{\delta}$ and $m^{*}_{obj, i}$\\
Let $LB \coloneqq m^{*}_{obj, i}$\\
Solve problem ~\eqref{Eqn:infeasible_MLD} with $\boldsymbol{\delta}$ using dual simplex\\
\If{Feasible}{
    Solve ~\eqref{Eqn:subproblem_MLD} with solutions from ~\eqref{Eqn:infeasible_MLD} as warm-starts\\
    Let the optimal cost be $f^{*}_{obj, i}$\\
    \If{$f^{*}_{obj, i}$<UB}
        {Let $UB \coloneqq f^{*}_{obj, i}$, $\boldsymbol{u}^{*} \coloneqq \boldsymbol{u}$\\}
    Add constraint ~\eqref{Eqn:optimal_cut_MLD} to master problem ~\eqref{Eqn:masterproblem_MLD_final}\\}
\Else(Infeasible)
    {Add constraint ~\eqref{Eqn:feasible_cutting_plane} to master problem ~\eqref{Eqn:masterproblem_MLD_final}\\
    \If{$\tilde{\boldsymbol\nu}_i^m$, $\tilde{\boldsymbol\lambda}_i^m$ is not equivalent to any existing $\tilde{\boldsymbol\nu}$, $\tilde{\boldsymbol\lambda}$} 
    {Add constraint
    ~\eqref{Eqn:feasible_cutting_plane_shifted} to master problem ~\eqref{Eqn:masterproblem_MLD_final}}}
$i \coloneqq i+1$\\}
\Return{$\boldsymbol{u}^*$}
\end{algorithm}

\section{Continual learning for warm-start}
\label{Sec:warm_start}
The idea behind the original Benders decomposition for linear programming subproblems \cite{benders1962partitioning} is to express the solutions to the dual of the subproblem by extreme rays and extreme points, and construct feasibility cuts and optimality cuts based on them. If all extreme rays and extreme points are found for the given subproblem, the master problem is guaranteed to find the global optimal $\boldsymbol{\delta}$. For the QP subproblem in this paper, the optimal solutions are not necessarily extreme points, but not far from them. In fact, $||\boldsymbol{A}^T\boldsymbol{\nu} + \boldsymbol{C}^T \boldsymbol{\lambda}||$ is minimized in ~\eqref{Eqn:feasible_dual}. If this norm is zero at the optimality, it is equivalent to the linear dual feasible set of ~\eqref{Eqn:Infeasible_dual}. As long as this norm is bounded, the dual solutions $(\boldsymbol{\nu}^*, \boldsymbol{\lambda}^*)$ come from a bounded (hence totally bounded) region around the extreme points of $\boldsymbol{A}^T\boldsymbol{\nu} + \boldsymbol{C}^T \boldsymbol{\lambda}=0$, which can be finitely covered by open balls $\textit{B}_{\epsilon}(\boldsymbol{\nu}^*, \boldsymbol{\lambda}^*)$. However, the number of extreme rays and covers is astronomical, hence an iterative procedure is used to only add necessary cuts in searching for optimal solutions. This avoids completely constructing the problem-solution mappings such as explicit MPC \cite{bemporad2002explicit}, especially when the matrices $\boldsymbol{A}$, $\boldsymbol{C}$ are undetermined before the solver begins (for example, the robot may have an unknown payload until it is hand over).

In this paper, we extend this idea to continual learning with dynamic environment represented by $\boldsymbol{\theta}$ shifting online. Only a small number of extreme rays and covers are added for a given $\boldsymbol{\theta}$. The solver continuously generates more rays and covers as $\boldsymbol{\theta}$ shifts. The new rays and covers are retained, while the duplicated ones are discarded. Once the solution is retained, it is never removed. The retained solutions provide increasingly better warm-starts for the incoming new problem instances. Since the number of extreme rays and covers is finite, this provess will terminate and the master problem does not grow infinitely large. This approach bares similarity of the continual learning framework\footnote[1]{A large number of literature continual learning is based on tasks (\cite{thrun1995learning, thrun1995lifelong, ruvolo2013ella}, to name a few). On the other hand, this work does not define tasks, hence more in-line with the task-free continual learning such as \cite{aljundi2019task}.}.

As MPC proceeds online, problem ~\eqref{Eqn:MLD} needs to be constantly resolved with different initial conditions $\boldsymbol{x}_{ini}$ and parameter $\boldsymbol{\theta}$. Since $\boldsymbol{x}_{ini}$ and $\boldsymbol{\theta}$ take the same position in the subproblem ~\eqref{Eqn:subproblem_MLD} as $\boldsymbol{\delta}$, all the optimality cuts ~\eqref{Eqn:optimal_cut_MLD} that construct lower bounds for $\boldsymbol{\delta}$ are also valid lower bounds for changing $\boldsymbol{x}_{ini}$ and $\boldsymbol{\theta}$. In addition, the feasibility cuts ~\eqref{Eqn:feasible_cutting_plane}
can also be used for new initial condition as $\tilde{\boldsymbol\nu}_{p}$, $\tilde{\boldsymbol\lambda}_{p}$ are independent of $\boldsymbol{x}_{ini}$ and $\boldsymbol{\theta}$. Assume we have feasible and optimality cuts as listed in ~\eqref{Eqn:masterproblem_MLD_final}. When new initial condition $\boldsymbol{x}_{ini}^{\prime}$ and different parameter $\boldsymbol{\theta}^{\prime}$ come in, we update the cutting planes:

\begin{equation}
\begin{aligned}
&for \ p=1, ..., \text{current \# of infeasible subproblem:} \\
&\boldsymbol{b}(\boldsymbol{x}_{ini}^{\prime}, \boldsymbol{\delta})^{T} \tilde{\boldsymbol\nu}_{p} + \boldsymbol{d}(\boldsymbol{\theta}^{\prime}, \boldsymbol{\delta})^{T} \tilde{\boldsymbol\lambda}_{p} \geq 0
\end{aligned}
\label{Eqn:infeasible_cut_updated}
\end{equation}

\begin{equation}
\begin{aligned}
for \ q=1, ..., &\text{current \# of optimal subproblem} \\
    z_0 \geq f_{obj,q}^* &+ \boldsymbol\nu_{q}^{*T} (\boldsymbol{b}(\boldsymbol{x}_{ini}, \boldsymbol{\delta}_q) - \boldsymbol{b}(\boldsymbol{x}_{ini}^{\prime}, \boldsymbol{\delta})) \\
    & + \boldsymbol\lambda_{q}^{*T} (\boldsymbol{d}(\boldsymbol{\theta}, \boldsymbol{\delta}_q)-\boldsymbol{d}(\boldsymbol{\theta}^{\prime}, \boldsymbol{\delta}))
\end{aligned}
\label{Eqn:optimal_cut_updated}
\end{equation}    

\begin{corollary}
For given $\boldsymbol{x}_{ini}^{\prime}$ and $\boldsymbol{\theta}^{\prime}$, any $\boldsymbol{\delta}$ that contradicts ~\eqref{Eqn:infeasible_cut_updated} proves infeasibility for ~\eqref{Eqn:infeasible_MLD}.
\end{corollary}
\begin{proof}
    A simple result from Lemma (\ref{Lemma2}) given $\tilde{\boldsymbol\nu}_{p}$ and $\tilde{\boldsymbol\lambda}_{p}$ are independent of $\boldsymbol{x}_{ini}$ and $\boldsymbol{\theta}$.
\end{proof}

\begin{corollary}
\begin{equation*}
    f_{obj,q}^* + \boldsymbol\nu_{q}^{*T} (\boldsymbol{b}(\boldsymbol{x}_{ini}, \boldsymbol{\delta}_q) - \boldsymbol{b}(\boldsymbol{x}_{ini}^{\prime}, \boldsymbol{\delta})) +
    \boldsymbol\lambda_{q}^{*T} (\boldsymbol{d}(\boldsymbol{\theta}, \boldsymbol{\delta}_q)-\boldsymbol{d}(\boldsymbol{\theta}^{\prime}, \boldsymbol{\delta}))
\end{equation*}
gives a lower bound of $v(\boldsymbol{x}_{ini}^{\prime}, \boldsymbol{\theta}^{\prime}, \boldsymbol{\delta})$.
\end{corollary}

\begin{proof}
A simple result from that $\boldsymbol{x}_{ini}$, $\boldsymbol{\theta}$ and $\boldsymbol{\delta}$ take the same position in ~\eqref{Eqn:optimal_cut_MLD}.
\end{proof}

With this technique, when a new initial condition is received, we run the updated master problem first. The master problem automatically provides a good warm-start using knowledge of previously accumulated cuts, reducing the number of iterations. The modified MPC algorithm with continual learning is provided in Algorithm \ref{Algorithm:with_ws}. 

\begin{algorithm}
\caption{Benders MPC with continual learning}
\label{Algorithm:with_ws}
\KwIn{$\boldsymbol{x}_{ini}, \ \boldsymbol{\theta}, \ G_a, \ \epsilon$}
\textit{Initialization} $LB \coloneqq -\infty$, $UB \coloneqq \infty$, iteration $i \coloneqq 0$\\
Update $\boldsymbol{x}_{ini}$, $\boldsymbol{\theta}$ in all the existing feasibility and optimality cuts in master problem ~\eqref{Eqn:masterproblem_MLD_final}\\
\While{$|UB-LB|/|UB| \geq G_a$}{
Solve master problem ~\eqref{Eqn:masterproblem_MLD_final} to get $\boldsymbol{\delta}$ and $m^{*}_{obj, i}$\\
Let $LB \coloneqq m^{*}_{obj, i}$\\
Solve problem ~\eqref{Eqn:infeasible_MLD} with $\boldsymbol{\delta}$ using dual simplex\\
\If{Feasible}{
    Solve ~\eqref{Eqn:subproblem_MLD} with solutions from ~\eqref{Eqn:infeasible_MLD} as warm-starts\\
    Let the optimal cost be $f^{*}_{obj, i}$, optimal dual variables be $\boldsymbol\nu_{i}^*, \boldsymbol\lambda_{i}^*$\\
    \If{$(\boldsymbol\nu_{i}^*, \boldsymbol\lambda_{i}^*) \notin \textit{B}_{\epsilon}(\boldsymbol{\nu}_i^*, \boldsymbol{\lambda}_i^*)$} 
    {Add constraint ~\eqref{Eqn:optimal_cut_MLD} to master problem ~\eqref{Eqn:masterproblem_MLD_final}\\
        \If{$f^{*}_{obj, i}$<UB}
            {Let $UB \coloneqq f^{*}_{obj, i}$, $\boldsymbol{u}^{*} \coloneqq \boldsymbol{u}$\\}}}
\Else(Infeasible)
    {Add constraint ~\eqref{Eqn:feasible_cutting_plane} to master problem ~\eqref{Eqn:masterproblem_MLD_final}\\
    \If{$\tilde{\boldsymbol\nu}_i^m$, $\tilde{\boldsymbol\lambda}_i^m$ is not equivalent to any existing $\tilde{\boldsymbol\nu}$, $\tilde{\boldsymbol\lambda}$} 
    {Add constraint
    ~\eqref{Eqn:feasible_cutting_plane_shifted} to master problem ~\eqref{Eqn:masterproblem_MLD_final}}}
$i \coloneqq i+1$\\}
\Return{$\boldsymbol{u}^*$}
\end{algorithm}




\section{Experiment}
\label{Sec:Experiment}

We test our Benders MPC to control the inverted pendulum with moving soft walls. This is also presented as a verification problem in previous works \cite{aydinoglu2023consensus, marcucci2020warm, cauligi2021coco}, except that we additionally randomize the wall motion. 

\begin{figure}[t!]
    \centering
    \includegraphics[width=0.45\textwidth]{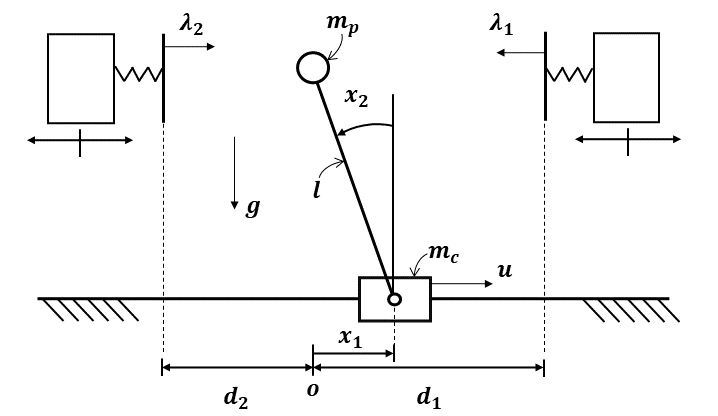}
     \caption{Cart-pole system with moving soft contact walls. \label{fig:inverted_pendulum}}
\end{figure} 

\subsubsection{Problem setup}
The setup is shown in Fig. \ref{fig:inverted_pendulum}. Let the nonlinear pendulum dynamics be $\dot{\boldsymbol{x}} = f(\boldsymbol{x}, \boldsymbol{u}) + \boldsymbol{n}$. $x_1$ is the position of the cart, $x_2$ is the angle of the pole, and $x_3$, $x_4$ are their derivatives. The control input $\boldsymbol{u}$ is the horizontal actuation force to push the cart. $\boldsymbol{n}$ is a random disturbance torque acting on the pole. The moving elastic pads are located to the right of the origin at a distance of $d_1$, and to the left at a distance of $d_2$. Let $l$ be the length of the pole. When the pole penetrates ($x_1-lcos(x_2) \geq d_1$ or $x_1-lcos(x_2) \leq -d_2$), additional contact force is generated at the tip of the pole. Let the parameter $\boldsymbol{\theta} = \matt{d_1 & d_2}$. 

We linearize the pendulum model around $x_2=0$ and use a linear elastic law for the wall contact. The linear model is:

\begin{equation}
\begin{aligned}
    \dot{x}_1 &= x_3 \\
    \dot{x}_2 &= x_4 \\
    \dot{x}_3 &= \frac{gm_p}{m_c}x_2 + \frac{u}{m_c}\\
    \dot{x}_4 &= \frac{g(m_c + m_p)}{lm_c} x_2 + \frac{u}{lm_c} + \frac{\lambda_1}{lm_p} - \frac{\lambda_2}{lm_p}\\
\end{aligned}
\end{equation}

Where $m_p$ is the mass of the pole, $m_c$ is the mass of the cart, $\lambda_1$, $\lambda_2$ are contact forces from the right and the left walls, respectively. They are both assumed to be positive. $g$ is the gravitational acceleration. We define the control input $\boldsymbol{u} = \matt{u & \lambda_1 & \lambda_2}^T$. If penetration happens, there is a non-zero contact force. This can be modeled as mixed-integer linear constraints:

\begin{equation}
\begin{aligned}
    \delta_i=0 &\Rightarrow \lambda_i = 0, \ a_i(lx_2 - x_1) + \frac{\lambda_i}{k_i} + d_i \geq 0 \\
    \delta_i=1 &\Rightarrow \lambda_i \geq 0, \ a_i(lx_2 - x_1) + \frac{\lambda_i}{k_i} + d_i = 0
\end{aligned}
\end{equation}

Where $i=1, 2$. $a_1=1$ and $a_2=-1$. $k_1$ and $k_2$ are elastic coefficients to model the right and left wall contacts. These logic laws are enforced as mixed-integer linear constraints using the standard big-M approach \cite{vielma2015mixed}, where the maximal distance possible from pole to wall, $D_{max}$, and maximal possible contact force, $\lambda_{max}$, are used as big-M constants. We also define the maximal and minimal cart position limits $d_{min}$ and $d_{max}$, and angle limits to be $\pm \frac{\pi}{2}$. The velocity limits of the cart, angular velocity limits of the pole, and control limits $u_{max}$ are also defined accordingly. This problem has $n_x=4$, $n_u=3$, $n_z=2$, $n_c=20$ (including variable limits). The matrices (variable limits are excluded) after discretization are defined such that:

\begin{equation}
    \boldsymbol{E} = \boldsymbol{I}_4 + dt\begin{bmatrix}
        0 & 0 & 1 & 0 \\
        0 & 0 & 0 & 1 \\
        0 & gm_p/m_c & 0 & 0 \\
        0 & g(m_c + m_p)/(lm_c) & 0 & 0 
    \end{bmatrix}
\end{equation}
\begin{equation}
    \boldsymbol{F} = dt\begin{bmatrix}
        0 & 0 & 0 \\
        0 & 0 & 0 \\
        1/m_c & 0 & 0 \\
        1/(lm_c) & 1/(lm_p) & -1/(lm_p)
    \end{bmatrix}
\end{equation}
\begin{equation}
    \boldsymbol{H}_1 = \begin{bmatrix}
        0 & 0 & 0 & 0 \\
        0 & 0 & 0 & 0 \\
        -1 & l & 0 & 0 \\
        1 & -l & 0 & 0 \\
        1 & -l & 0 & 0 \\
        -1 & l & 0 & 0
    \end{bmatrix} \
    \boldsymbol{H}_2 = \begin{bmatrix}
        0 & 1 & 0 \\
        0 & 0 & 1 \\
        0 & 1/k_1 & 0 \\
        0 & -1/k_1 & 0 \\
        0 & 0 & 1/k_2 \\
        0 & 0 & -1/k_2
    \end{bmatrix}
\end{equation}
\begin{equation}
    \boldsymbol{H}_3 = \begin{bmatrix}
        -\lambda_{max} & 0 \\
        0 & -\lambda_{max} \\
        D_{max} & 0 \\
        0 & 0 \\
        0 & D_{max} \\
        0 & 0 
    \end{bmatrix} \
    \boldsymbol{h}(\boldsymbol{\theta}) = \begin{bmatrix}
        0 \\
        0 \\
        -d_1 + D_{max} \\
        d_1 \\
        -d_2 + D_{max} \\
        d_2
    \end{bmatrix}
\end{equation}

Other matrices are zeros. The objective function penalizes the control efforts, the velocities, and tries to regulate the pole to the zero position. We choose $\boldsymbol{Q}_k=\boldsymbol{I}_4$, $\boldsymbol{R}_k=\boldsymbol{I}_2$. The terminal cost $\boldsymbol{Q}_N$ is obtained by solving a discrete algebraic Ricatti equation. In the actual experiment, we choose $m_c=1.0kg$, $m_p=0.4kg$, $l=0.6m$, $k_1=k_2=50N/m$, $u_{max}=20N$. The discretization step size $dt=0.02s$ and planning horizon $N=10$. 

\subsubsection{Monte-Carlo experiment}

We implement Algorithm \ref{Algorithm:with_ws} to solve this problem. We choose $gap = 0.1$, which is identical among all the benchmark methods. The $\epsilon$ is chosen properly to reduce the number of optimality cuts. We use off-the-shelf solver Gurobi to solve both the master problems (MIPs) and the subproblems (QPs).



The controller is coded in Python, and tested inside a pybullet environment \cite{coumans2016pybullet} on a 12th Gen Intel Core i7-12800H × 20 laptop with 16GB memory. At the beginning of each test episode, the pendulum begins from a perturbed angle of $x_2=10 \degree$ such that it will bump into the wall to regain balance. For the rest of each episode, the persistent random disturbance torque $\boldsymbol{n}$ is generated from a Gaussian distribution with zero mean and a standard deviation $\boldsymbol{\sigma}=8Nm$. The system is constantly disturbed and has to frequently touch the wall for re-balance. The wall motion is generated by a random walk on top of a base sinusoidal motion with a constant offset $d_{\text{off},i}$:

\begin{equation}
    d_i = d_{\text{off},i} + Asin(wt+\theta_1) + m_i, \ i=1,2
\end{equation}



Where $m_i$ is the integration of a Gaussian noise. We conduct statistical analysis for 10 feasible trajectories under different disturbance torque $\boldsymbol{n}$ and wall motion $m_i$. The data is collected from solved problems where at least one contact is planned, removing the cases when contact is not involved. The disturbance torque is unknown to the controller. The wall motions are provide to controller only at run-time. 


The following methods are also used for benchmark:

\begin{enumerate}[label=(\roman*)]
    \item Warm-started Branch and Bound. We implemented the Branch and Bound algorithm in Python with warm-start as described by \cite{marcucci2020warm}.
    \item Off-the-shelf solver. We implemented the off-the-shelf solver Gurobi. The problem is setup only once and solved iteratively such that warm-starts are automatically used to minimize the solving time. The default setting is used to optimize the performance.
    \item GBD without warm-start. We implemented Algorithm \ref{Initial_algorithm} such that previous cuts are not used to warm-start the next problem.
\end{enumerate}

\subsubsection{Results}
\begin{figure}[t!]
    \centering
    \includegraphics[width=0.5\textwidth]{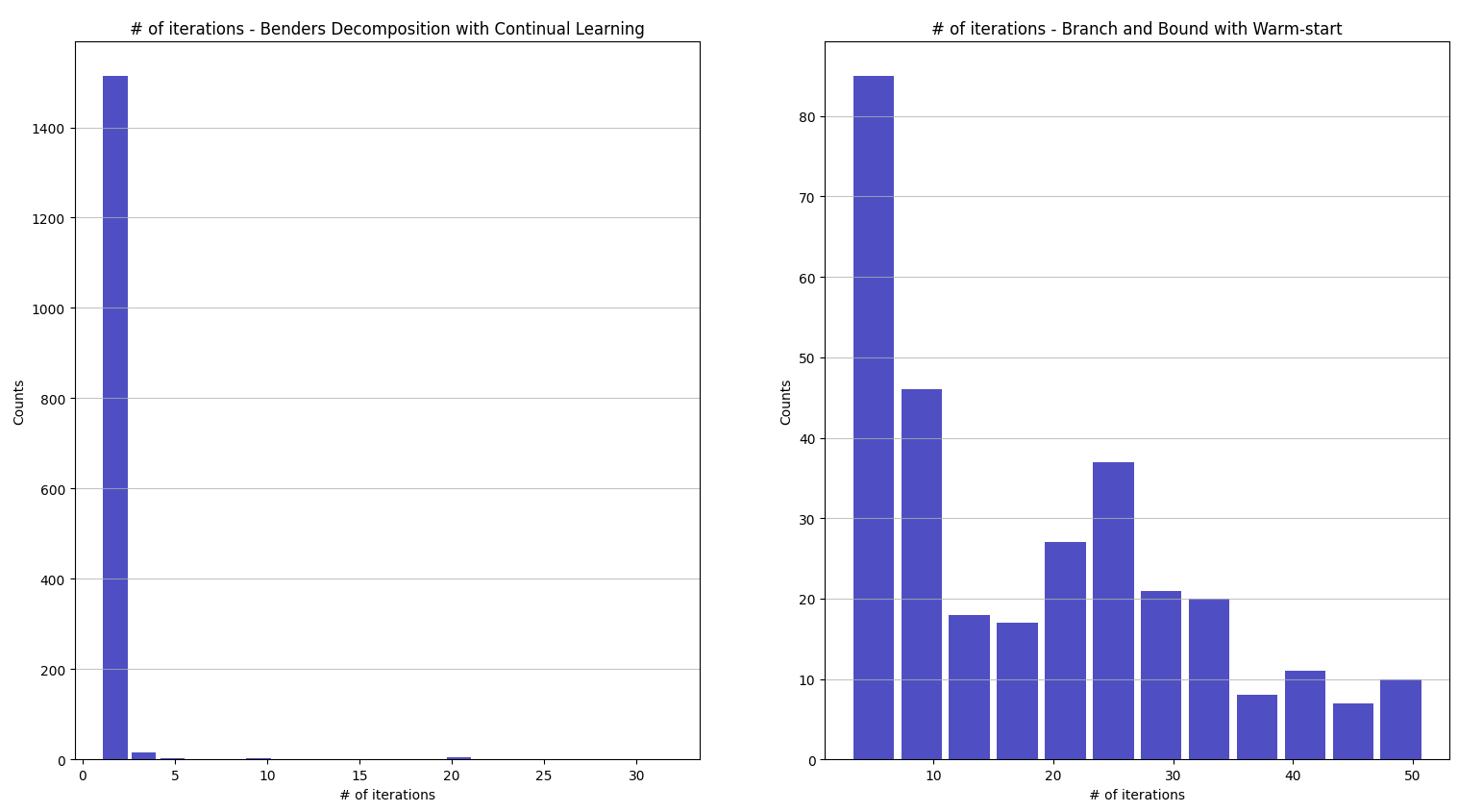}
     \caption{Comparison of number of solver iterations for different problems of $(\boldsymbol{x}_{ini}, \boldsymbol{\theta})$. x-axis is the range of solver iterations. y axis is the count of problem instances from the collected trajectories. Left: The proposed GBD with continual learning. Right: Branch and Bound with warm-start \cite{marcucci2020warm}.}
     \label{fig:Avg_solving_speed-pendulum}
\end{figure}

Fig. \ref{fig:Avg_solving_speed-pendulum} gives the histogram result showing the number of iterations to solve the problem and their frequencies. Thanks to the continual learning, 99.2\% of problem instances are solved within 5 iterations by GBD, except for a few problems during the cold-start phase taking more than 10 iterations. This 99.2\% problem instances have an average solving speed of $500-600Hz$. This solving time represents a complete procedure of Algorithm \ref{Algorithm:with_ws}, where the time spend in solving master and subproblems account for 60\% of the total time. Note that previous work \cite{rahmaniani2017benders} reported over 90\% solving time spent on the master problem. On the contrary, we report that the master problem is oftentimes solved within the presolve stage, since hundreds of cuts are accumulated after a few problem instances. This accounts for less than 30\% of the total solving time.

On the other hand, Branch and Bound algorithm relies on subproblems. The warm-start scheme reduces the number of solved subproblems by 50\%. However, the BB solver still goes through more than 10$\times$ subproblems to converge compared to the GBD solver, from our averaged data. The gist of this warm-start scheme is to shift the covers in time. Ideally, the computations up till $k=N-2$ can be reused and the algorithm only needs to compute the binary input at $k=N-1$. However, this scheme becomes less effective as the amplitude of wall motion $A$ increases. The reason is that the current contact sequence cannot simply be shifted and has to be recomputed before $k=N-1$. Consequently, additional covers need to be refined.

\begin{figure}[t!]
    \centering
    \includegraphics[width=0.5\textwidth]{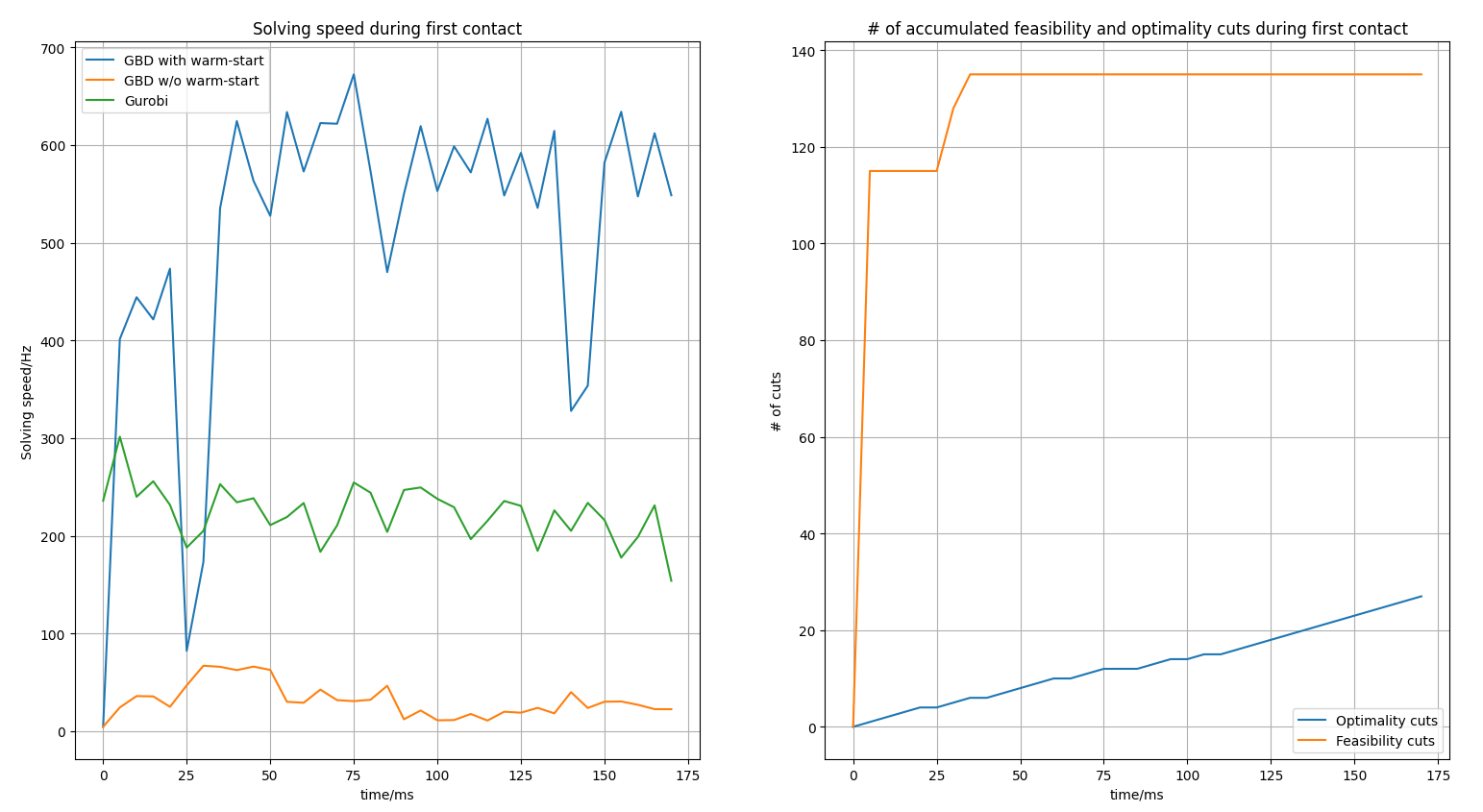}
     \caption{A case of solving procedure from cold-start when the pole bumps into the moving elastic wall. x-axis is time and y-axis is the solving speed in Hz. Left: Comparison of solving speed between GBD with continual learning for warm-start, GBD without any warm-start, and off-the-shelf solver Gurobi. Right: The number of cuts accumulated during the solving procedure.}
     \label{fig:contact_analysis}
\end{figure} 

Fig. \ref{fig:contact_analysis} shows the solving speed in Hz during the beginning of an episode from our data. The solver begins from cold-start but has to plan contact ever since $t=0$. After one iteration of from cold-start (taking $200ms$ in Fig. \ref{fig:contact_analysis}), the cuts accumulate to provide warm-start for the next iteration, and solving speed increases over Gurobi ($200-300Hz$). Without warm-start, the solving speed remains on average $25Hz$.

Due to the fast cold-start, even if the system dynamics are only known at run-time or completely change, the time cost to learn new dynamics for our problem is at the scale of hundreds of milliseconds. This is much faster than training neural-network-based policies \cite{cauligi2021coco}. If global optimal solutions are not required in the beginning, the learning time can be further reduced.



\section{Conclusion, Discussion and Future Work} 
\label{Sec:conclusion}
In this paper, we proposed a hybrid MPC solver based on Generalized Benders decomposition with continual learning. The algorithm accumulates cutting planes from the invariant dual space of the subproblems under than randomly changing environment. After a cold-start phase at the scale of hundreds of milliseconds, the accumulated cuts provide warm-starts for the new problem instances to increase the solving speed. This leads to solving speeds that are 2-3 times faster than the commercialized solver Gurobi in controlling the cart-pole system with randomly moving soft walls.

There are several theoretical analyses and hardware experiments that can make this preliminary results more thorough. For example, analyzing the generalizability of the learned cuts to the new problem instances, or testing the scalability of this algorithm to more complex problems. We can also combine Branch and Bound with Benders cuts to leverage both of their strengths. Although we already have results to some of the questions above, they do not fit into the current paper and will be included in the future journal version.

\begin{appendices}


\section{Proof of lemma 1}
\label{Appendix_pf_lemma1}
We present a proof of Lemma \ref{Lem:lemma1}. Recall the Farkas' lemma (Theorem 4.6 in \cite{bertsimas1997introduction}):

\begin{theorem}
\label{Thm:alternative}
    Let $\tilde{\boldsymbol{A}} \in \mathbb{R}^{m \times n}$ and $\tilde{\boldsymbol{b}} \in \mathbb{R}^{m}$. Then, exactly one of the two following alternatives holds:
    \begin{enumerate}
        \item There exists some $\tilde{\boldsymbol{x}} \geq \boldsymbol{0}$ such that $\tilde{\boldsymbol{A}} \tilde{\boldsymbol{x}} = \tilde{\boldsymbol{b}}$.
        \item There exists some vector $\tilde{\boldsymbol{p}}$ such that $\tilde{\boldsymbol{p}}^T \tilde{\boldsymbol{A}} \geq \boldsymbol{0}^T$ and $\tilde{\boldsymbol{p}}^T \tilde{\boldsymbol{b}} < 0$.
    \end{enumerate}
\end{theorem}

For any vector $\boldsymbol{x}$, there exists $\boldsymbol{y} \geq \boldsymbol{0}$, $\boldsymbol{z} \geq \boldsymbol{0}$ such that $\boldsymbol{x} = \boldsymbol{y} - \boldsymbol{z}$. The inequality constraint $\boldsymbol{C} \boldsymbol{x} \leq \boldsymbol{d}$ is equivalent to $\boldsymbol{C} \boldsymbol{x} + \boldsymbol{\delta} = \boldsymbol{d}, \exists \boldsymbol{\delta} \geq \boldsymbol{0}$. Hence the first condition of Lemma \ref{Lem:lemma1} is equivalent to the existence of $\tilde{\boldsymbol{x}} = \matt{\boldsymbol{y}^T & \boldsymbol{z}^T & \boldsymbol{\delta}^T}^T \geq 0$ such that:

\begin{equation}
    \underbrace{\begin{bmatrix}
        \boldsymbol{A} & -\boldsymbol{A} & \boldsymbol{0} \\
        \boldsymbol{C} & -\boldsymbol{C} & \boldsymbol{I}
    \end{bmatrix}}_{\tilde{\boldsymbol{A}}} 
    \underbrace{\begin{bmatrix}
        \boldsymbol{y} \\
        \boldsymbol{z} \\
        \boldsymbol{\delta}
    \end{bmatrix}}_{\tilde{\boldsymbol{x}}}
    = \underbrace{\begin{bmatrix}
        \boldsymbol{b} \\
        \boldsymbol{d}
    \end{bmatrix}}_{\tilde{\boldsymbol{b}}}
\end{equation}

By Theorem \ref{Thm:alternative}, this condition is alternative to the existence of $\tilde{\boldsymbol{p}} = \matt{\boldsymbol{y}^T & \boldsymbol{z}^T}^T$ such that $\tilde{\boldsymbol{p}}^T \tilde{\boldsymbol{A}} \geq \boldsymbol{0}^T$ and $\tilde{\boldsymbol{p}}^T \tilde{\boldsymbol{b}} < 0$, which gives the second condition of Lemma \ref{Lem:lemma1}.



\end{appendices}

\textbf{\textit{Acknowledgements}}
The author would like to thank Zehui Lu, Shaoshuai Mou, and Yan Gu for their helpful discussions and suggestions.


{
\bibliographystyle{IEEEtran}
\bibliography{references}
}


\end{document}